\documentclass[10pt,twocolumn,letterpaper]{article}

\usepackage{cvpr}
\usepackage{times}
\usepackage{epsfig}
\usepackage{graphicx}
\usepackage{amsmath}
\usepackage{amssymb}
\usepackage{algorithm}
\usepackage{multirow}
\usepackage{algorithmic}
\usepackage{mathrsfs}
\usepackage{mathtools}
\usepackage{subfigure}
\usepackage{enumitem}
\usepackage{tensor}
\usepackage{xcolor}
\usepackage{booktabs}
\usepackage{balance}
\usepackage{caption}
\usepackage[normalem]{ulem}
\algsetup{linenosize=\small}
\pdfoutput=1
\usepackage[pagebackref=true,breaklinks=true,letterpaper=true,colorlinks,bookmarks=false]{hyperref}

\cvprfinalcopy % *** Uncomment this line for the final submission

\def\cvprPaperID{3905} % *** Enter the CVPR Paper ID here

\begin{document}

%%%%%%%%% TITLE
\title{TBNet:Pulmonary Tuberculosis Diagnosing System using Deep Neural Networks}

\author{Ram Srivatsav Ghorakavi\\
Australian National University, Canberra, ACT\\
{\tt\small ram.ghorakavi@anu.edu.au}
%\and
%Suryansh Kumar\\
%Australian National University\\
%Canberra, ACT\\
%{\tt\small suryansh.kumar@anu.edu.au}
}

\twocolumn[{%
\renewcommand\twocolumn[1][]{#1}%
\maketitle
\begin{center}
    \centering
    \includegraphics[width=0.90\textwidth,height=6.8cm]{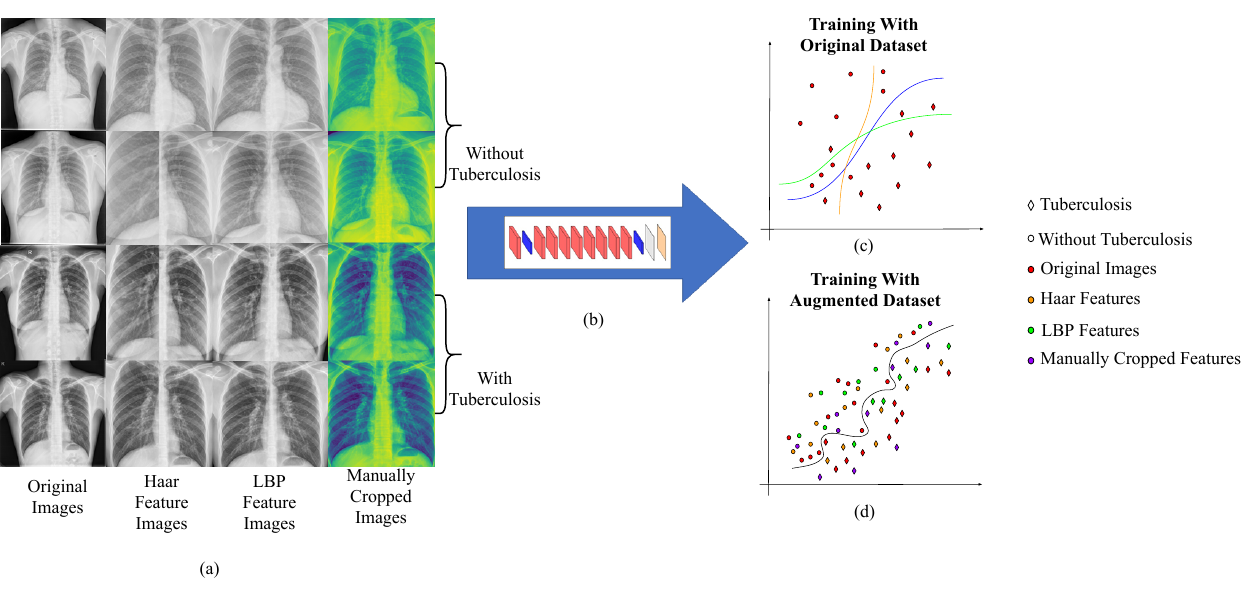}
    \captionof{figure}{(a) Data Augmentation using salient features. In addition to original image, Haar-features and LBP features are used to enhance the inference of tuberculosis. (b) Res-Net18 Architecture (c) Graph shows that using only original image features is not able to generalize the classification, hence, multiple decision boundaries are shown. (d) By using the same small set of original data I can generate informative salient features and by employing data augmentation a reliable decision boundary can be achieve. The procured decision boundary provides a more accurate generalization for tuberculosis diagnosis.}
    \label{fig:firstPage}
\end{center}%
}]

%%%%%%%%% ABSTRACT
\begin{abstract}
Tuberculosis is a deadly infectious disease prevalent around the world. Due to the lack of proper technology in place, the early detection of this disease is unattainable. Also, the available methods to detect Tuberculosis is not up-to a commendable standards due to their dependency on unnecessary features, this make such technology obsolete for a reliable health-care technology. In this paper, I propose a deep-learning based system which diagnoses tuberculosis based on the important features in Chest X-rays along with original chest X-rays. Employing our system will accelerate the process of tuberculosis diagnosis by overcoming the need to perform the time-consuming sputum-based testing method (Diagnostic Microbiology). In contrast to the previous methods \cite{kant2018towards, melendez2016automated}, our work utilizes the state-of-the-art ResNet \cite{he2016deep} with proper data augmentation using traditional robust features like Haar \cite{viola2005detecting,viola2001rapid} and LBP \cite{ojala1994performance,ojala1996comparative}. I observed that such a procedure enhances the rate of tuberculosis detection to a highly satisfactory level. Our work uses the publicly available pulmonary chest X-ray dataset to train our network \cite{jaeger2014two}. Nevertheless, the publicly available dataset is very small and is inadequate to achieve the best accuracy. To overcome this issue I have devised an intuitive feature based data augmentation pipeline. Our approach shall help the deep neural network \cite{lecun2015deep,he2016deep,krizhevsky2012imagenet} to focus its training on tuberculosis affected regions making it more robust and accurate, when compared to other conventional methods that use procedures like mirroring and rotation. By using our simple yet powerful techniques, I observed a 10\% boost in performance accuracy.

% Tuberculosis is an infectious disease which is transmitted through the air and can result in death. The good news is that this disease can be curable when caught in its early stages. Tuberculosis has a high rate of occurrence in countries like India, China, Indonesia and South Africa \cite{}. Diagnosing Tuberculosis disease is a time-consuming job requiring a diagnosis from both the doctor and a Radiologist. With this project, we designed a system which diagnoses a person based on just the chest X-rays. Introducing this system in real time can prevent the time consumed in diagnosing a person and allow for large-scale testing. In this project, we are using a convolution neural network to perform the task of diagnosing by training it on the pulmonary chest X-Ray dataset available on Kaggle. This system gives suggestions about the diagnosis which can help doctors in diagnosing a person faster.
\end{abstract}

%%%%%%%%% BODY TEXT
\section{Introduction}
% Tuberculosis is a deadly infectious disease which is transmitted when a person comes in contact with the bacteria. The transfer of this disease can occur via air, water or any other form of contact with bacteria medium, avoiding these mediums is really hard as they are common day to day utilities. Due to its nature, this disease was widely spread all around the world making it one of the top ten mortal diseases. Many first world countries has taken initiative in preventing the spread of this disease by employing high health care standards and technology. The employed diagnosing standards and technology of this disease usually deal with human intervention, as a result they are time consuming and highly prone to errors. Looking into currently available prolonged testing methods, they have high chance of occurance of faulty test outcome due to human involvement, resulting in misdiagnosis of the patients. Tuberculosis is curable when caught early, but because of aforementioned problems, diagnosing a person can be delayed and resulting in incurable state of tuberculosis.
Tuberculosis (TB) is a deadly communicable disease. This disease can spread via air, water or any other forms of day-to-day activities. As a result, its one of the widely spread illness all around the world, making it one of the deadly disease.   To prevent the spread of this disease many countries have taken serious initiative by employing high health care standards and technology. However, the standards and technology employed to diagnose this disease usually need human interventions, as a result, they are costly, time-consuming and highly prone to errors. Such prolonged testing methods have a high chance of faulty outcome, which can result in mis-diagnosis of the patients. In medicine, it's well known that Tuberculosis is curable when caught at an early stage, but due to such an obsolete procedure in place, it difficult for doctors to diagnose TB on time, resulting in an incurable state of tuberculosis.

% Usually a person with tuberculosis will show the symptoms like cough, chest pains, weakness, weight loss, fever and night sweats. These symptoms serve as a primary indication of the disease requesting the person for immediate hospital admittance. In hospitals, diagnosing TB involves the chest radiography and sputum smear testing. The chest X-Rays are used to check the abnormalities in the lung regions indicating the signs of tuberculosis. However, the abnormalities observed in the chest X-Rays do not unequivocally indicate that a person has tuberculosis, these abnormalities can indicate any other disease causing the similar abnormalities in the chest X-rays. So, doctors perform sputum test that is done to confirm the inference from the chest X-ray. In this test sputum excreted by the infected is observed for the bacteria responsible for tuberculosis called acid-fast bacilli bacteria under microscope. This method is very effective in diagnosing a person with TB but culturing a medium of testing is time-consuming and will take at least 24 hours to make it viable for diagnosis. All these tests require human involvement to observe or to perform the test.

Diagnosing TB involves the ``chest radiography'' and ``sputum smear'' testing. The ``chest radiography'' is used to check the abnormalities in the lung regions indicating the signs of tuberculosis. However, other abnormalities caused due to cough, chest pains, \etc observed in the chest X-rays do not unequivocally indicate that a person has tuberculosis. As a result, doctors perform an extra sputum test to confirm the inference from the chest X-ray. In this test sputum excreted by the infection is observed for the bacteria responsible for tuberculosis called ``acid-fast bacilli" bacteria under a microscope. This method is very effective in diagnosing a person with TB but preparing the appropriate medium for this test takes at least 24 hours to make it viable for diagnosis, which is time consuming. Additionally, all these tests require diligent human involvement to perform the test.

% As humans dependent testing methods are highly subjective to errors, we need a means to autonomously perform tests. Addressing these testing autonomous will speed-up the process of testing resulting in a fast and accurate diagnosis of a person. Through autonomous testing we can just confine ourselves to single testing method either to sputum test or using a chest X-rays. As autonomous testing mechanisms are designed to observe minute changes in the object of interest, we can utilise only one method of testing. As sputum testing method is more accurate, employing a autonomous sputum based tuberculosis system can generate very accurate results, but sputum based testing method is very time consuming as the test initially involves the culture period of at least half a day which makes sputum ready for diagnosis. Few of the tuberculosis diagnosis technology, are designed based on the sputum based tuberculosis detection method, but they weren’t able to reduce the time taken to culture the observable medium. Because of this problem we need an alternative method which is fast and less time consuming to address this issue. So, The best way is via chest X-ray based tuberculosis detection.

As humans dependent testing methods are subject to errors, we need a high precision autonomous system to perform this task. As alluded to above, the testing method can be performed using sputum or X-ray. Although sputum-based tuberculosis system can generate very accurate results, it takes at least \textbf{a day} to complete the TB diagnosis.  Therefore, we need an alternative method which is accurate, fast and consumes less time to address this issue.  Leveraging on the success of modern deep neural networks on image detection and recognition \cite{he2016deep,krizhevsky2012imagenet,lecun2015deep}, I utilized  X-ray data to supply a fast and reliable system for TB detection.

It is well-studied in deep learning that deep neural networks requires a large number of training data samples to perform well \cite{krizhevsky2012imagenet}. However in medical imaging, collecting large dataset is not easy as it requires individual consent. Consequently, previous approaches use Support Vector Machine (SVM)\cite{cortes1995support} based algorithms on publicly available small dataset to perform the tuberculosis detection, hence, not scalable. In contrast, in this work I propose a new data augmentation technique which helps deep neural network to work more efficiently on limited TB X-ray images. Our data augmentation technique conveniently utilizes traditional fast Haar features \cite{viola2001rapid}, LBP feature \cite{ojala1994performance}, and cropping of region of interest to improve the performance of TB detection.

In this paper, I have focused on developing a deep neural-network based algorithm for diagnosing tuberculosis using chest X-rays. For this purpose, I have used the pulmonary chest X-rays of 800 different people taken from the dataset created by National Library of Medicine, Maryland, the USA. the dataset is composed of the data collected from Shenzhen hospital and Montgomery hospital. The dataset collected from Shenzhen hospital has 662 chest X-ray images, and the Montgomery hospital has 138 chest X-ray images. This dataset is available as a free source for the educational and research purpose. I used both of these datasets to train a deep neural network (ResNet). Our work makes the following contributions:
\begin{enumerate}[nolistsep, itemindent=0em, label=(\alph*)]
\item I present a method that can be used to diagnose a person for tuberculosis based on the pulmonary chest X-rays.
\item In our method, I used the traditional features like Haar, LBP and cropped region of interest to perform data augmentation, which improves the performance of the deep neural network by making it more focused on tuning itself with the important features in the chest X-rays.
\item I have also studied the effect of increase in testing accuracy on the network when the incomplete feature data is added in data augmentation step.
\end{enumerate}

\section{Related Work and Motivation}
% Looking into recent research, there are many Machine learning and computer-vision based tuberculosis detection algorithms which are employed in diagnosing a person with tuberculosis. Of them, there are two approaches are very significant. One way is using the microscopic images of sputum and perform the feature detection or machine learning approaches to detect the presence of tuberculosis. The other way is using chest radiograph, where we utilise the feature detection or machine learning approaches to identify the abnormalities caused in lungs due to tuberculosis. Initial research focused on designing tuberculosis detection system using microscopic images of the sputum. In this method, the image processing algorithms utilised are focused on identifying and isolating and detecting the bacteria causing tuberculosis called Acid-fast-bacilli (AFB) from its surroundings. Diagnosing a person with this method sometimes might not be very accurate as the bacteria can be misleading resulting in false positive, moreover developing the culture to visualise the bacteria in the microscope is time-consuming. Despite being accurate the sputum based approach was employed for tuberculosis detection in the real world scenarios.

There are many computer-vision and machine-learning based tuberculosis detection algorithms \cite{melendez2016automated, lakhani2017deep, veropoulos1998image,Veropoulos1999AutomatedIO}. Nevertheless, there are two prominent ways to solve this problem. One way is to use the {\bf{microscopic images}} of sputum to perform the feature detection and use machine learning based approaches \cite{cortes1995support} to detect the presence of tuberculosis. The other way is to use {\bf{chest radiograph}} image features and use machine learning approaches to identify the abnormalities caused in the lungs due to tuberculosis \cite{lakhani2017deep}.

Initial research focused on designing tuberculosis detection system uses {\bf{microscopic images}} of the sputum \cite{veropoulos1998automated, Veropoulos1999AutomatedIO, veropoulos1998image}. In such methods, the image processing algorithms were utilized to focus on identifying, isolating and detecting the bacteria causing tuberculosis called Acid-fast-bacilli (AFB) from its surroundings.  Diagnosing a person with this method sometimes may not be very accurate as the bacteria can be misleading, resulting in false positive. Moreover, developing the culture to visualize the bacteria in the microscope is time-consuming. Despite such limitations, the sputum-based approach is used in practice for tuberculosis detection due to its par accuracy. 

% The second method is the utilisation of chest X-rays to diagnosis the disease. In this method we utilise the x-rays to observe the abnormalities in the lung region. In this approach, we can use many image processing and machine learning techniques to identify and isolate the abnormal regions in the lungs. As this method of diagnosing a person based on X-rays is not very accurate because of its similarities with other diseases. we need a method which can keep track of minute differences that can be used to differentiate tuberculosis from other diseases. As a result using an image-processing based approach to address this problem will not generate accurate results. So, we focused on learning based approach in diagnosing the disease with chest X-rays.

The second kind of methods use chest X-rays images to observe the abnormalities in the lung regions. These methods apply several image processing techniques to identify and isolate the abnormal regions in the lungs. As these methods for TB detection is based on X-rays image processing algorithm which supplies similar response for other diseases too, hence, it's not reliable and accurate. Therefore, we need a method which can keep track of minute differences that can be used to differentiate tuberculosis from other diseases. As a result, relying solely on image-processing based approaches to address this problem is not recommended.  This motivates to design an algorithm that learns the salient features from the X-ray images which can reliably supply information about tuberculosis. 
% us to utilize deep learning  along with image features approach in diagnosing the disease with chest X-rays.

% In this paper, we have employed supervised approach in training algorithm to detect the tuberculosis. Here the output diagnostic results of the input chest X-ray images are previously defined and the algorithm is trained based on the diagnostic results and the input chest X-ray images. In the learning based approach, the best algorithm that can be employed for small datasets is SVM. SVM is widely recognised method which work very well for the cases with small training dataset. This feature of SVM can be quite beneficial for medical applications as the publicly available datasets are very small. In this scenario, application of neural network will not result in better accuracy. In this paper, we devised a method which improves the accuracy of the neural network and makes it perform as good as SVM.
In this paper, I have applied a supervised approach to detect tuberculosis.  Our algorithm is trained based on the doctor's diagnostic results on the chest X-ray images. Since the publicly available datasets are very small, support vector machine algorithm's are often used to solve this problem. However, I show that by cleverly exploiting the same small scale dataset, I can make the large-scale data-driven deep-neural network work as good as SVM. This makes our algorithm applicable to large-scale tuberculosis detection systems and flexible for further usage as more and more dataset may get publicly available in the future.

% In recent year there were many attempts that are made in designing a system which can be used to diagnose tuberculosis, of them two research work have motivated us to pursue a new approach of achieving better accuracy and faster convergence via neural networks. The first research work is done by Jaime Melendez under the title ‘An automated tuberculosis screening strategy combining X-ray-based computer-aided detection and clinical information’ and the second work done by Paras Lakhani under the title ‘Deep Learning at Chest Radiography: Automated Classification of Pulmonary Tuberculosis by Using Convolutional Neural Networks’. The first approach designs a feature based diagnostic system which utilises computer aided detection (CAD) tool for identifying and isolation the infected tuberculosis infected region in the chest X-rays. In the second approach, they have developed a deep neural network to which is trained on the above discussed chest X-ray dataset along with the data collected from the hospitals Belarus TB Public Health Program and Thomas Jefferson University Hospital which are not publicly available as research and educational purposes. This has created an added advantage for this work to generate the best accuracy of 99\%. Now let us look into the step involved in both the methods.

Recently numerous attempt has been made to design a tuberculosis diagnosing system. Out of these approaches, a couple of recent research motivated us to pursue this work to achieve faster and accurate solution to this problem. Firstly, Melendez \etal \cite{melendez2016automated} proposed a feature based diagnostic system which utilizes computer-aided detection (CAD) tool to identify and isolate the infected tuberculosis regions in the chest X-rays. Secondly, the method proposed by Lakhani \etal \cite{lakhani2017deep} train deep neural network both on the publicly available dataset, and data collected from the hospitals Belarus TB Public Health Program and Thomas Jefferson University Hospital, which are not publicly available for research purpose. This has created an added advantage for this work to generate the best accuracy of 99\%.  Now, let's delve into the step involved in both the methods for clear understanding.

% \textbf{(a) An automated tuberculosis screening strategy combining X-ray-based computer-aided detection and clinical information [8]:} In this method they have developed a computer aided detection (CAD) tool which screens through the X-ray and clinical information to diagnose a person for tuberculosis.
% This method uses the dataset collected at Gugulethu TB Clinic (Cape Town, South Africa). Based on the x-ray data the significant features that contribute in tuberculosis are identified by using the CAD tool and are score along with the clinical information and performs a scoring algorithm based on the extent of the individual feature scores.This method was devised to overcome the drawback when the detection process is done with individual features. This method was very effective and has achieved an accuracy of {\bf{84\%}}.
\textbf{(a) Melendez \etal approach} \cite{melendez2016automated}: 
This method proposed a framework that uses the combination of Computer-aided detection (CAD) score automatically computed from a chest radiograph reading (CXR) and 12 clinical x-ray features to achieve high accuracy for TB screening. This method was devised to overcome the drawbacks of radiological interpretation systems in TB-endemic countries. This method uses the dataset collected at Gugulethu TB Clinic (Cape Town, South Africa) claims to achieves an effective accuracy of 84\%.

\textbf{(b) Lakhani \etal approach} \cite{lakhani2017deep}:
% In this method, they have used the deep convolutional neural network to diagnose a person based on the chest radiographs. In this approach they have employed Deep-ConvNet, AlexNet and GoogleLeNet to train their dataset. The dataset that they used is generated by collecting the data from four different hospitals to create a large pool of chest X-ray images. Before training they have applied conventional data augmentation methods like 90°, 180° and 270°  rotation and Contrast Limited Adaptive Histogram Equalization to further increase the volume of available dataset. They found that the pre-trained neural networks has generated the best accuracy when compared with untrained networks. Over that, they were able to achieve the best classification accuracy with GoogLeNet which is {\bf{98\%}} over 120 epochs for the case of pre-trained network with data augmentation.
This method uses a deep convolutional neural network to diagnose a person based on the chest radiographs. They employ Deep-ConvNet 
\cite{lecun2015deep}, AlexNet \cite{krizhevsky2012imagenet}, and GoogLeNet \cite{szegedy2015going} to train these network on their dataset. The dataset that they used is generated by collecting the data from four different hospitals to create a large pool of chest X-ray images. Before training, they have applied conventional data augmentation methods like 90°, 180°, and 270°  rotation and Contrast Limited Adaptive Histogram Equalization to further increase the volume of the available dataset.  Using this augmented dataset and pre-trained  GoogLeNet, they were able to achieve the classification accuracy of 98\% using GoogLeNet over 120 epochs.

% Inspired from both the works, we have devised a new method which can be employed to significantly reduce the training time by increasing the training loss convergence rate using feature salient extraction methods like Haar and LBP along with manual cropping. We all know training a convolutional neural networks like GoogLeNet and AlexNet for 120 epochs will result in overfitting of the data, the accuracy that the second work achieved might not be consistent and cannot be trusted in the real world scenarios. To make the system more robust and fast converging we suggest an add-on technique which involves Haar and LBP feature extraction methods along with manual cropping in the data augmentation process, which increases the focus of training a neural network on the salient tuberculosis features. For the experimentation purpose we have confined over selves with the data augmentation performed with  Haar and LBP feature extraction methods along with manual cropping methods, doing this will give us a clear idea on how this approach of data augmentation will improve the performance of the above discussed methods.

In this paper, I argue that the accuracy claimed by Lakhani \etal \cite{lakhani2017deep} may not generalize to real-world TB diagnosis system due to over-reliability on self-made dataset. Also, CAD detection tools are expensive and require additional hardware interfacing. In contrast to the these methods, I devised a new method which significantly reduces the training time by increasing the training loss convergence rate using salient features like Haar and LBP. It's well-known that training a convolutional neural network like GoogLeNet and AlexNet for 120 epochs results in over-fitting of the data. To make the system more robust and fast, I suggest an add-on technique which involves Haar features, LBP feature extraction methods with additional focus on region of interest via cropping lungs regions, in the data augmentation process. Such procedure increases the focus of training a neural network on the salient tuberculosis features. Our approach of data augmentation based on salient features improves both the performance and response time of TB diagnosis system, even with a small number of data samples in hand.

\begin{figure*}[t] 
\begin{center}
\includegraphics[width=0.90\linewidth, height=7.0cm]{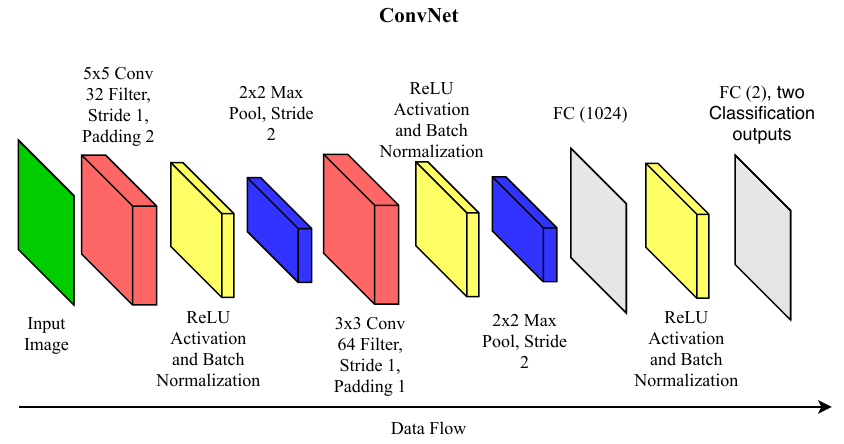}
\caption{Deep-ConvNet Architecture. (\textbf{Green}) Input Image (\textbf{Light Pink}) Conv Layer with 1 stride and 2 padding (\textbf{Yellow}) Rectified Linear Unit (\textbf{Blue}) Max pool Layer (\textbf{Gray}) Fully Connected Layers.\label{fig:networkconv}}
\end{center}
\end{figure*}

\section{Deep Learning based classification model}\label{ss:mldl}
Inspired from the recent work \cite{melendez2016automated,lakhani2017deep}, I developed a method which is designed to improve the accuracy of tuberculosis detection system. Our system uses deep neural network with an efficient way of data augmentation. To generate an accuracy that can be applied in the real-world application with the deep neural networks, we need at least 20,000 x-ray images, but publicly available dataset is composed of just 800 images. So, I need a data augmentation method to make our system more reliable and accurate. Our approach of data augmentation significantly improves the performance of the neural network. I first perform few experiments to decide which network can predict tuberculosis with reasonable precision on original dataset. For this, I use two networks, first network is composed of two convolution layers and two fully connected layers and the other is pre-trained ResNet18. The accuracy generated by ResNet18 when trained on original dataset ({\bf{65.77181\%}}) which is significantly higher when compared to the first network ({\bf{52.34899\%}}). As a result, I used our base network as the pre-trained ResNet18 to perform experiments.

In our approach to increase the accuracy of ResNet18, I have performed the data augmentation process using feature like Haar \cite{viola2001rapid}, LBP \cite{ojala1994performance} and cropped region of interest. These process generates the data of salient features which can be added to the pre-existing dataset of 800 images. It is well-known that large number of dataset can make the performance of deep neural network better than traditional machine learning algorithms like gradient descent, SVM \etc. Using such notions, I augmented the salient feature with original images to improve the accuracy of tuberculosis diagnosing system. This operation of adding salient features with original images directs the network  to focus on important information of the chest X-ray images resulting in significant and fast convergence of the network. Before going into details of the data augmentation process, let us investigate how feature extraction methods work.

\textbf{(a) Haar Feature Extraction:} This method is widely used in extracting features from a digital image to perform object recognition operation. Each object in nature has features which makes it stand out in the image. One way to extract these features is by comparing the intensity values of them with its neighbors. This operation is done pixel wise by considering each pixel to be a independent feature. In this algorithm, I perform the averaging operation on the pixel intensities surrounding the pixel of interest and use it to compare with the required average pattern of the average pixel intensity values of the reference lung image. The reference average pixel intensity values of an image are nothing but the predefined Haar features of lungs. The output of this algorithm is discussed in results section.

\textbf{(b) LBP (Local Binary Patterns) Feature extraction:} In this method, I perform the similar operation as Haar feature extraction method, but instead of averaging the neighboring pixel intensity values, I compare the intensity value of the interested pixel with its neighboring pixel intensities and assign a bit pattern to each neighboring pixel via comparison results. Later this bit pattern in the window of interest is compared with the reference bit pattern of the object of interest (lungs) to identify lungs. The reference bit pattern of the object of interest is nothing but predefined LBP features of lungs. The output of this algorithm is discussed in results section.

Both the methods were very effective in isolation the lung region from the chest X-rays. The accuracy of the Haar feature extraction method is {\bf{71.125\%}} where 569 lung regions out of 800 are correctly segmented. Whereas in the LBP feature extraction method the accuracy of salient feature extraction is {\bf{79.125\%}} where 633 lung regions out of 800 are correctly isolated. Later I have cropped the region of interest from the original dataset to increase the robustness of our data augmented method. The incorrectly identified lung regions generated in LBP feature extraction were grouped separately and are called noisy data further in this paper. Finally, to test the new approach to data augment, I have created three cases based approach. The description of these cases is as follows.

\begin{itemize}
\item {\bf{Case 1:}} Data augmentation with one feature extraction Algorithm. (No noisy data added).
\item {\bf{Case 2:}} Data augmentation with two feature extraction Algorithm and cropping region of interest (Noisy data generated from LBP feature extraction is added).
\item {\bf{Case 3:}} Data augmentation with two feature extraction Algorithm and cropping region of interest (No noisy data added).
\end{itemize}

In case 1 I have added features generated using Haar feature extraction along with the original dataset to check if this approach increases the accuracy. As the above test is successful, I have tested the case 2 and 3, where I performed the same operation as case 1 but instead of just one feature extraction method added data from all the three salient feature extraction methods. The difference in case 2 and 3 is that I have added the noisy data generated by the LBP feature extraction method is added in case 2 to test if the noisy data increases the performance of the neural network. Our idea was if the data generated by feature extraction method has at least a few features related to the region of interest, considering this data in training process should increase the accuracy of the network. This theory was proven to be accurate by generating the highest accuracy in the case 2 over the other cases. The detailed discussion of these results are done in the section (\ref{ss:resutls}).
\begin{figure*}[t] 
\begin{center}
\includegraphics[width=1.0\linewidth, height=7.0cm]{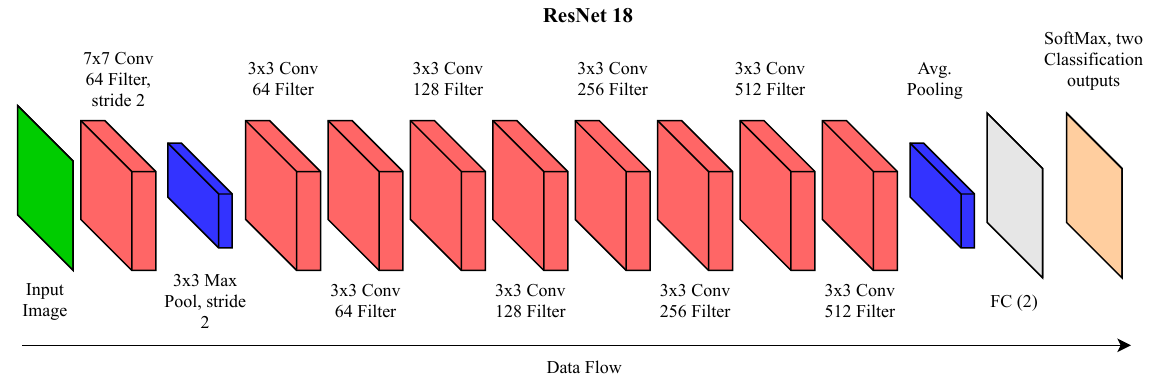}
\caption{ResNet18 Architecture \cite{he2016deep}. (\textbf{Green}) Input Images, (\textbf{Pink})Convolution Layer, (\textbf{Blue}) Pooling Layer, (\textbf{Gray}) Fully-Connected Layer (\textbf{Apricot}) SoftMax. \label{fig:networkres18}}
\end{center}
\end{figure*}

I used widely employed cross entropy loss-function to train both Conv-Net and ResNet 18. Let $\hat{y_n}$ and $y_n$ be the predicted and supplied probability of having tuberculosis respectively. I defined our cross-entropy as

\begin{equation}
\begin{aligned}
& \displaystyle L = -\frac{1}{N_{a}}\sum_{i=1}^{N_a} \Big[y_nlog(\hat{y_n}) + (1-y_n)log(1- \hat{y_n}) \Big]
\end{aligned}\label{eq:1}
\end{equation}
Here, `$N_a$' denoted the total number of samples after data augmentation. The weights of the loss are updated using ADAM optimization \cite{kingma2014adam} \ie
\begin{equation}
\begin{aligned}
& \displaystyle m_{w}^{t+1} \leftarrow \beta_1m_{w}^{t} + (1-\beta_1)\nabla_wL^{t}\\
& \displaystyle v_{w}^{t+1} \leftarrow \beta_2v_{w}^{t} + (1-\beta_2)(\nabla_wL^{t})^2 \\
& \displaystyle \hat{m}_w := \frac{m_w^{t+1}}{1-\beta_1^{t+1}} \\
& \displaystyle \hat{v}_w := \frac{v_w^{t+1}}{1-\beta_2^{t+1}} \\
& \displaystyle w^{t+1} \leftarrow w^{t} - \eta \frac{\hat{m}_w}{\sqrt{\hat{v}_w} + \epsilon}
\end{aligned}\label{eq:2}
\end{equation}

where $m, v$ are the first and second moment vector respectively. $\beta_1$ and $\beta_2$ are the forgetting factors for gradients and second moments of gradient. $\epsilon$ is a small scalar value to avoid division by zero.
The overall procedure of our method is presented in Algorithm-1.
\begin{algorithm}[h!]
\label{Algorithm}
\caption{: \textbf{TBNet}}
\begin{algorithmic}
\STATE {\bf{Input:}} Augmented Dataset obtained using Haar, LBP features and cropped region of interest.
\STATE {\bf{Output:}} Diagnosis result of tuberculosis. 
 \STATE 1: Salient feature Extraction.
 \begin{itemize}[noitemsep]
 \item Haar Features extraction.
 \item LBP Features extraction.
 \item Cropping region of interest.
 \end{itemize}
 \STATE 2: Augmentation with Haar, LBP features and Cropped region of interest over original dataset.
\STATE 3: Splitting Augmented dataset :- test, train and validation set splitting based on the configuration shown in Table (\ref{tab:networktable}) and Table (\ref{tab:networktable2})
\STATE 4: Tuning pre-trained ResNet18 using the Cross Entropy Loss Eq.(\ref{eq:1}) and Adam optimizer Eq.(\ref{eq:2}).
\STATE 5: Diagnosis result.
\end{algorithmic}
\end{algorithm}

\section{Experiments and Results}\label{ss:resutls}
To simulate this model, I have used computing system with 16GB RAM, 6GB GTX1060 GPU with CUDA 9.2 interfacing. Our algorithm is simulated on python3.6. Our implementation can be also be realized on commodity desktop machine. Our experimentation is performed in two steps: In the first step I observed the effects of training Deep-ConvNet and ResNet18 when trained on original dataset. The first network \ie Deep-ConvNet, is composed of two convolution layers and two fully connected layers which are tuned to generate the classification output during the training process see Fig.(\ref{fig:networkconv}). The detailed specifications of the Deep-ConvNet are listed below in order of execution.

\begin{itemize}
\item $5\times5$ Convolutional Layer with 32 filters, stride 1 and padding 2.
\item ReLU Activation Layer
\item Batch Normalization Layer
\item 2x2 Max Pooling Layer with a stride of 2
\item 3x3 Convolutional Layer with 64 filters, stride 1 and padding 1.
\item ReLU Activation Layer
\item Batch Normalization Layer
\item 2x2 Max Pooling Layer with a stride of 2
\item Fully-connected layer with 1024 output units
\item ReLU Activation Layer
\item Fully-connected layer with 2 output unit
\end{itemize}

\begin{figure*}[!htp]
  \begin{center}
  \subfigure[\label{fig:plot1}]{\includegraphics[width=0.32\linewidth , height=0.25\linewidth]{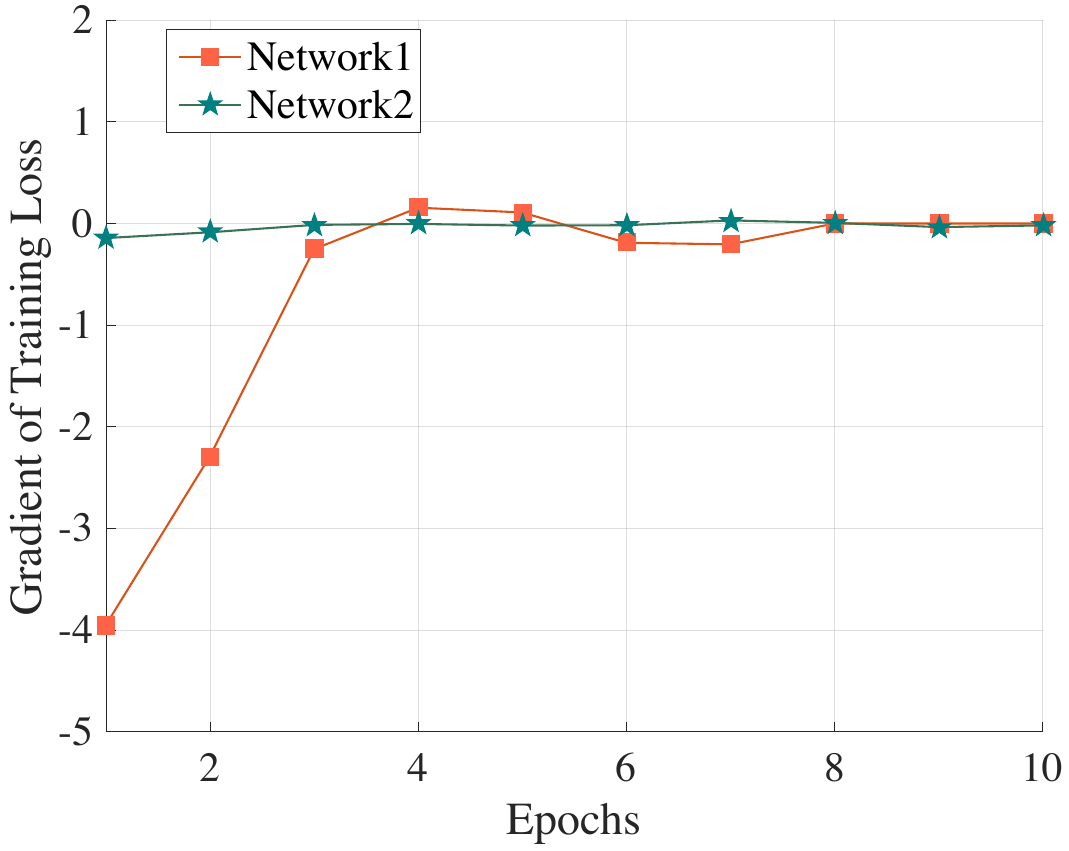}}
  \subfigure[\label{fig:plot2}]{\includegraphics[width=0.32\linewidth, height=0.25\linewidth]{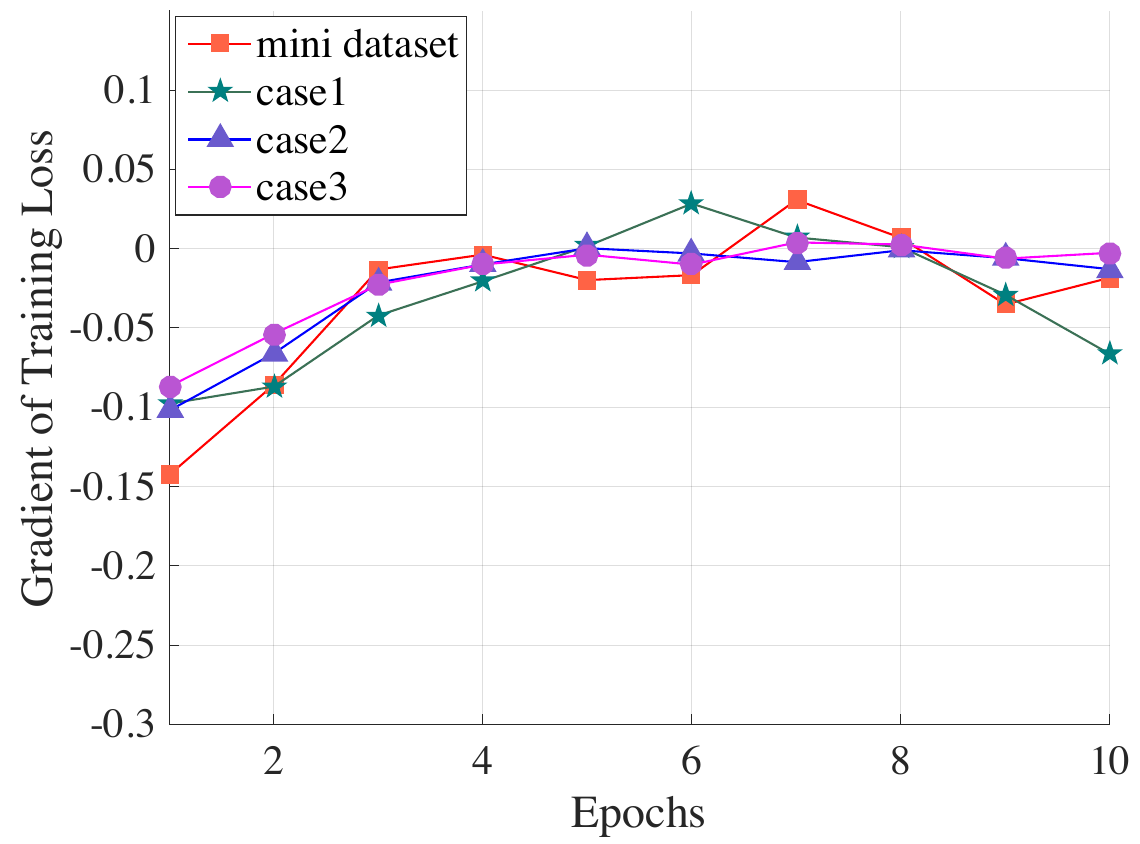}}
  \subfigure[\label{fig:plot3}]{\includegraphics[width=0.32\linewidth, height=0.25\linewidth]{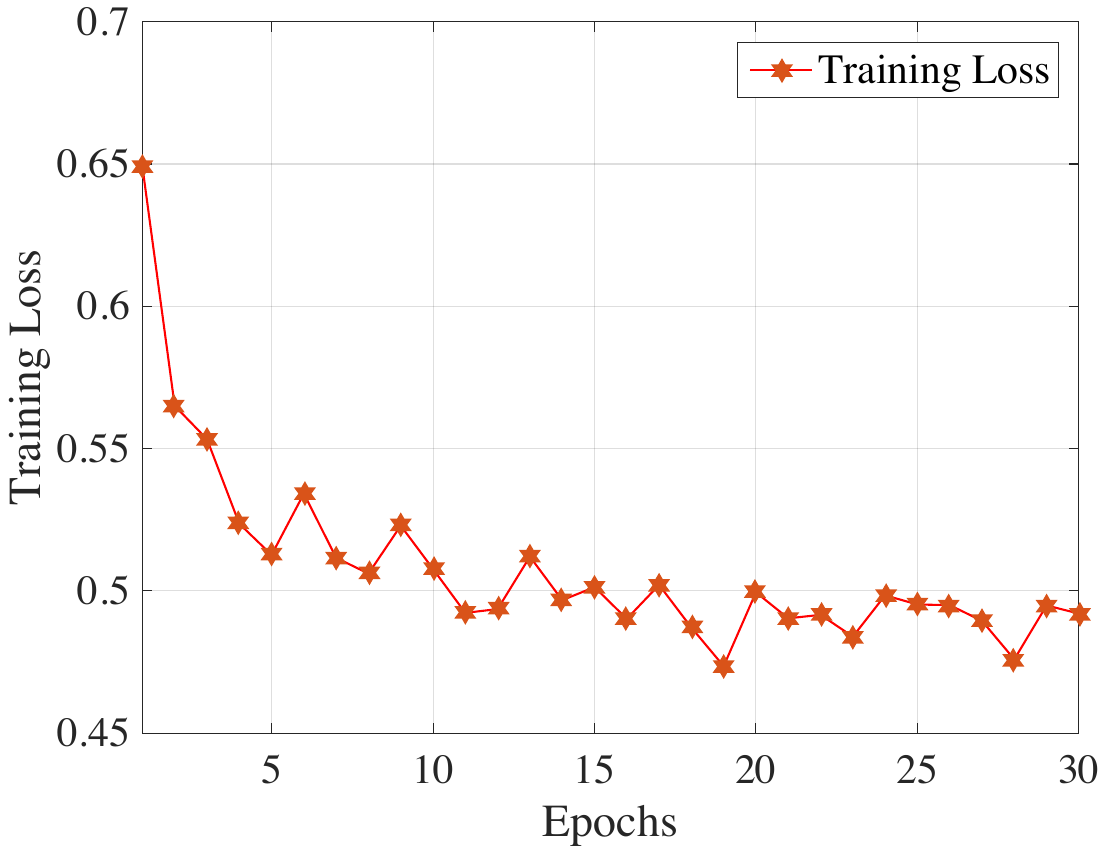}}
\end{center}
  \caption{\small (a).Training loss Convergence of Network1 (Deep-ConvNet \textbf{Fig.}~(\ref{fig:networkconv})) and Network2 (ResNet18 \textbf{Fig.}~(\ref{fig:networkres18})) (b). Rate of Convergence of training loss on all three cases discussed in $\S \ref{ss:mldl}$ along with network 2 trained on original dataset (c). Case 2 $\S \ref{ss:mldl}$ training loss graph for 30 epochs.}
  \label{fig:statisticalPlot}
\end{figure*}

Later I have tuned the pre-trained ResNet18 on original data to compare its performance with Deep-ConvNet. The detailed architecture of ResNet 18 can be observed in the Fig.(\ref{fig:networkres18}). As the network parameters in Deep-ConvNet are less when compared with the ResNet18 this Increase will create more room for the network to train, resulting in better accuracy when compared with Deep-ConvNet. Sometimes parameterized network might result in over fitting, with this experiment I decided the best network that can be utilized in the later stages. By fine Tuning the pretrained ResNet 18 network I have achieved better testing accuracy when tested on 150 images ({\bf{65.77181\%}} testing accuracy) ---see Table (\ref{tab:networktable}). Over that, the rate of convergence of the training loss is much smoother and faster when compared with Deep-ConvNet ---see Fig.(\ref{fig:plot1}). As a result, I decided ResNet18 as the base network to perform the experimentation in the later stages. Looking into the training parameters, all the experiments were run on a Batch size of 50 and Epoch size of 10. As there is no geometrical model defining the abnormalities occurred due to tuberculosis in the lung regions caused due to TB, the training process of ResNet18 is done in a supervised fashion.

\begin{table}
\small
\centering
\begin{tabular}{c|c|c|c}
\hline
Neural Network        & Training Size & Testing Size & \%Accuracy\\ \hline
Network1 & 650   & 150    &  58.34899\%  \\ \hline
Network2 & 650   & 150    &  {\bf{65.77181}}\%  \\ \hline
\end{tabular}
\caption{ Percentage accuracy of tuberculosis detection using Network1 (Deep-ConvNet) and Network2(ResNet18) when trained on original data without augmentation.}\label{tab:networktable}
\end{table}

\begin{figure}[t] 
\begin{center}
\includegraphics[width=1.0\linewidth]{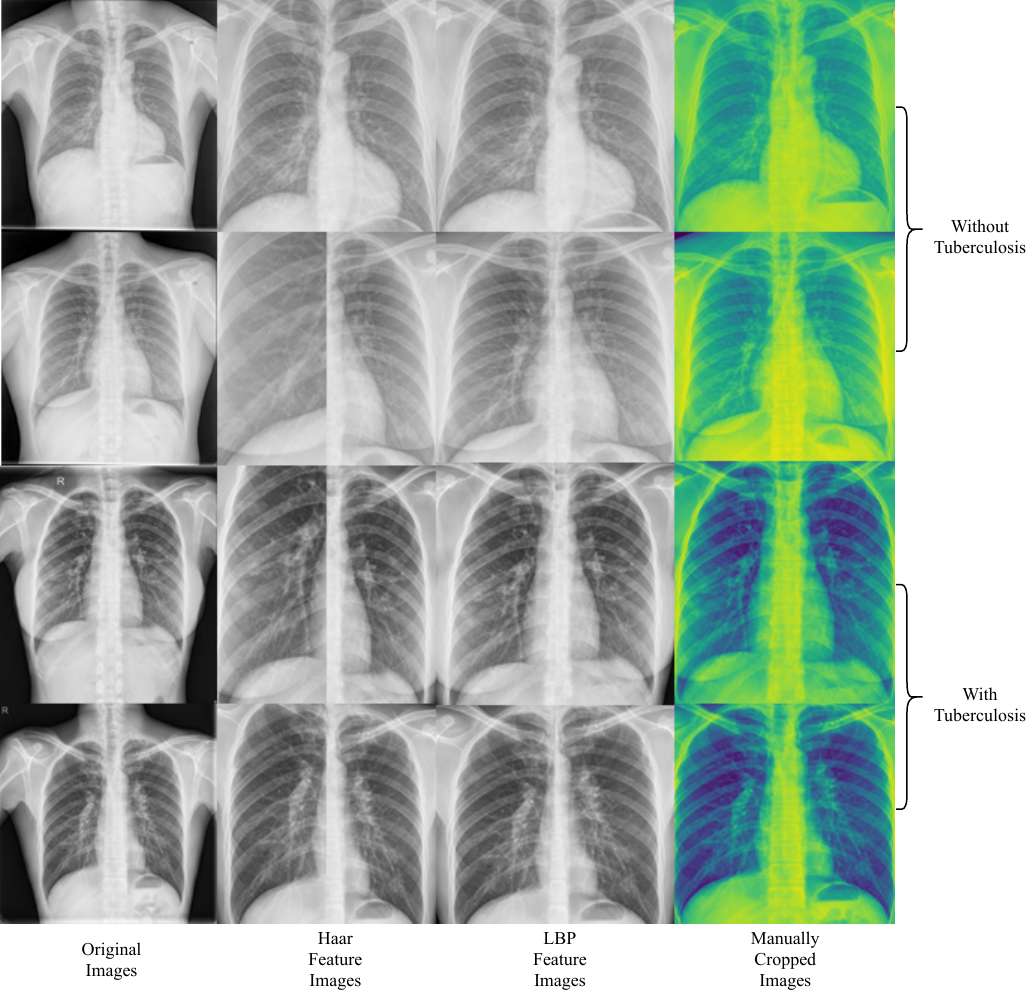}
\caption{From left to right Raw data, Haar data, LBP data and manually cropped data.\label{fig:picture1}}
\end{center}
\end{figure}

After observing the testing accuracy of the above experiment ---see Fig.(\ref{fig:plot1}),Table (\ref{tab:networktable}), I found that the accuracy achieved by the pre-trained network can be further improved. So, I have performed data augmentation using feature extraction algorithms to extract the salient features from the original dataset instead of conventional data augmentation approaches used in deep-learning. The figure \ref{fig:picture1} shows the raw data, Haar features, LBP features and cropped region of interest. All the data generated is pooled together and used to perform the training operation on the ResNet 18 network.

\begin{figure}[t] 
\begin{center}
\includegraphics[width=1.0\linewidth]{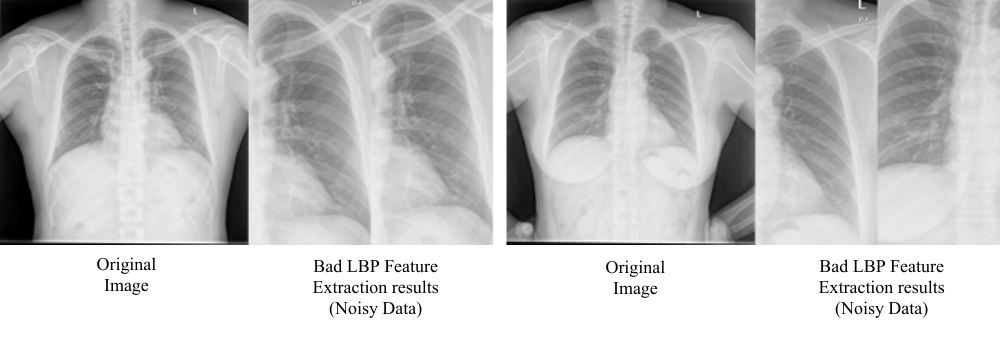}
\caption{From left to right, Original image, Bad Haar feature extraction result (noisy data), Original image, bad LBP feature extraction result (noisy data)\label{fig:picture2}}
\end{center}
\end{figure}

The data extracted using LBP feature extraction is more accurate than the Haar feature extraction. The LBP feature extraction has the accuracy of {\bf{79.125\%}} in identifying the features whereas the Haar feature extraction has the accuracy of {\bf{71.125\%}}. The output from the feature extractors is shown in the figure (\ref{fig:picture1}). While performing the feature extraction both the processes generate noisy data as in the figure (\ref{fig:picture2}) which is due to the similarities in the both the left and right lung regions. For example from the first couple of images it is evident that the similarities of left lung with right lung resulted in one lung being repeated twice. Sometimes, the lungs were also reversed in position as in the second image couple.

After extraction of the salient features, the pooling of data is done in three cases to clearly observe the effects of data augmentation using feature extraction methods. In the first case, I add the data generated using Haar feature extraction to the original data. In the second case I have added the data generated from both Haar and LBP feature images with original dataset along with the manually cropped salient feature images. I also added noisy image data generated during LBP feature extraction method to test if partially extracted features can improve the accuracy of the testing accuracy of the network. Finally in the Third case, I have added the data generated from both Haar and LBP feature extractions with original dataset along with the manually cropped salient feature extraction. These data cannot be ignored as they have few features related to TB. So, these data are pooled together and are called noisy data.

\begin{table*}
\centering
\begin{tabular}{c|c|c|c|c|c|c}
Database      & \begin{tabular}[c]{@{}l@{}}Dataset size\\~ ~(\#images)\end{tabular} & Training set & Validation set & Testing set & \begin{tabular}[c]{@{}l@{}}Validation Accuracy\\~ ~ (10th epoch)\end{tabular} & Testing Accuracy  \\ \hline
Original data & 800                     & 650          & 100            & 150         & 70.0\%                                                        & 65.77181\%        \\ \hline
Case 1        & 1,379                   & 1000         & 100            & 379         & 64.0\%                                                        & 72.48322\%        \\ \hline
Case 2        & 2,969                   & 2557         & 150            & 412         & {\bf{81.333}}\%                                                      & {\bf{75.42579}}\%        \\ \hline
Case 3        & 2,812                   & 2400         & 150            & 412         & 78.667\%                                                     & 74.69586\% \\ \hline        
\end{tabular}
\caption{ \small{Case1: Performance using original image and Harr features. Case3: Performance using Harr, LBP without noisy features. Case2:  Performance using Harr, LBP with noisy features. By observing the statistics of Case 2, it can observed that by using salient features like Harr, LBP along with noisy features can help boost the performance of the deep neural network.}}\label{tab:networktable2}
\end{table*}

The results of the three cases are discussed in the Table (\ref{tab:networktable2}). By comparing the testing accuracy of case 1 and case 3, the data augmentation performed with multiple feature extraction algorithms has generated more accuracy when compared with a single machine learning algorithm. Over that, when I compare the case 2 and case 3, the case 2 with noise data addition has generated best accuracy when compared with case 3, hence proving our idea, discussed in the section (\ref{ss:mldl}). Moreover, the rate of convergence is case 2 is much smoother and faster than the other cases, indicating the increase in performance over other approaches in data augmentation, as shown in the Fig.(\ref{fig:plot2}).

\section{Limitations}
From the above results, I observed that the accuracy of the network is much better in case 2 with multiple salient feature based data augmentation with original dataset. But if the network is trained for more epochs (30 epochs) the optimization rate of training loss is oscillating around a fixed point after 15th epoch (see Fig.\ref{fig:plot3}). This occurred because of the network parameters have reached their optimal point. To improve the accuracy further we can add additional convolution layers which result in increase in number of parameters which can be used to improve the accuracy of the system.

Addition of the convolution layers to ResNet 18 is not advisable because the newly added layers have to be trained from scratch resulting in increased training time. To prevent this I would suggest to utilize the higher ResNet versions like ResNet20 which have more  pre-trained convolution layers when compared with ResNet18. In this paper, I am not exploring this aspect as the argument that I am trying to make is different. If the accuracy is still not increased with the addition of convolution layers, addition of conventional data augmentation methods like horizontal flipping etc., with the proposed data augmentation will definitely increase the accuracy. Resulting in the case with the best possible accuracy achievable with the available data.

\section{Conclusion and Future Work}
% Tuberculosis is a deadly disease and by assisting the doctors with developing a diagnosing system which identifies this disease will save a lot of time invested in diagnosing. The problem of diagnosing tuberculosis using machine learning approach is a prevailing problem all around the world with very little recognisable research performed. The main reason for this is due to very limited data available for research and educational purposes. In this paper, we have new approach to tackle this problem by applying data augmentation with the help machine learning algorithm by making use of the available dataset. We have proved that salient features based data augmentation achieved using machine learning algorithms can be quite effective in improving the performance of the current state of the art system. A more accurate version of this could be deployed in developing countries where thousands of patients can be diagnosed in seconds. Our approach to addressing this problem by fusing feature extraction methods and deep learning has been proven effective in increasing accuracy and performance. Implementation of our method in the current diagnostic systems can improve their efficiency. This system would help humanitarian efforts done by an organisation like the WHO in relation with tuberculosis by saving time and money. 

I witnessed that by proper data augmentation techniques, we can make the state-of-the-art deep neural network work very efficiently on a small sample of available chest X-ray image dataset. In this paper, I have also demonstrated that salient features like Haar and LBP can be quite effective in improving the performance of the current state-of-the-art algorithms. The proposed approach in addressing this problem by fusing feature extraction methods and deep learning model has shown a substantial increase in the performance accuracy of the tuberculosis diagnosis system in very few epochs. Hence, such a system can be put to work on a real-time platform shall contribute in early diagnosis of TB. Lastly, I believe this algorithm can help save time, effort and money for health care organizations invested in treating TB. Furthermore, as my future work, I hope to employ 3D chest X-ray data to obtain structure from motion of lungs and achieve efficient TB abnormalities detection. The recent work done by Kumar \etal  \cite{kumar2016multi, kumar2017spatio, kumar2017monocular, kumar2018scalable, kumar2019jumping, kumar2019motion} look very promising and can help improve the TB detection using 3D chest X-ray data.

{\small
\bibliographystyle{ieee}
\bibliography{egbib_final}
}

\end{document}